\documentclass{article}

\usepackage{arxiv}
\pdfoutput=1

\usepackage[utf8]{inputenc} 
\usepackage[T1]{fontenc}    
\usepackage{hyperref}       
\hypersetup{colorlinks=true, linkcolor=blue, citecolor=blue, urlcolor=black}
\usepackage{url}            
\usepackage{booktabs}       
\usepackage{amsfonts}       
\usepackage{nicefrac}       
\usepackage{microtype}      
\usepackage{lipsum}		
\usepackage{graphicx}
\usepackage[numbers,square,sort&compress]{natbib}
\usepackage{doi}
\usepackage{graphicx}%
\usepackage{float}
\usepackage{xcolor}
\usepackage{algorithm}
\usepackage{algpseudocode}
\usepackage{multirow}%
 \usepackage{placeins}
\usepackage{amsmath,amssymb,amsfonts}%
\usepackage{amsthm}%
\usepackage{mathrsfs}%
\usepackage[title]{appendix}%
\usepackage{xcolor}%
\usepackage{textcomp}%
\usepackage{manyfoot}%
\usepackage{booktabs}%
\usepackage{algorithm}%
\usepackage{algorithmicx}%
\usepackage{algpseudocode}%
\usepackage{listings}%
\usepackage{placeins}
\usepackage{enumitem}%
\usepackage{tikz} 
\usetikzlibrary{arrows.meta, positioning} 
\usepackage{longtable}
\usepackage{amsthm}           
\newtheorem{lemma}{Lemma}    

\usepackage{amsthm}

\theoremstyle{plain}
\newtheorem{theorem}{Theorem}
\newtheorem{proposition}[theorem]{Proposition}

\theoremstyle{definition}

\theoremstyle{remark}
\newtheorem{remark}{Remark}

\title{Selected-Risk Calibration for Feasibility at Returned Decisions}

\date{} 					

\author{ \href{https://orcid.org/0009-0009-0866-8674}{\includegraphics[scale=0.06]{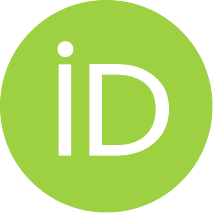}\hspace{1mm}Raghav Thind} \\
	Department of Computer Science\\
	University of Maryland, College Park\\
	\texttt{rthind@terpmail.umd.edu} \\
	\And
	Youran Sun \\
	Department of Mathematics\\
	University of Maryland, College Park\\
	\texttt{sun1245@umd.edu} \\
	\And
	Ling Liang \\
	Department of Mathematics\\
	University of Tennessee, Knoxville\\
	\texttt{liang.ling@u.nus.edu} \\
	\And
	Haizhao Yang \\
	Department of Mathematics\\
	Department of Computer Science\\
	University of Maryland, College Park \\
	\texttt{hzyang@umd.edu} \\
}

\renewcommand{\shorttitle}{Selected-Risk Calibration}

\hypersetup{
pdftitle={A template for the arxiv style},
pdfsubject={q-bio.NC, q-bio.QM},
pdfauthor={David S.~Hippocampus, Elias D.~Striatum},
pdfkeywords={First keyword, Second keyword, More},
}

\begin{document}
\maketitle

\begin{abstract}
	When an optimization problem uses a surrogate constraint, a decision that is feasible for the surrogate need not be feasible for the true constraint. We propose Selected-Risk Calibration (SRC), a framework that calibrates a learned constraint against the optimizer’s returned decisions for objectives drawn from a given distribution. SRC guarantees that, with probability at least $1-\delta$, every margin it certifies on held-out objectives has selected risk, the probability that the returned decision is infeasible, of at most $\gamma$. We also show that the random-input coverage guarantee of split conformal prediction does not bound the selected risk, and that likelihood-ratio reweighting for feedback covariate shift does not apply when the learned constraint is active at the returned decision. Our experiments show that a conformal margin covering about $90\%$ of random inputs covered as few as $17\%$ of the returned decisions, and every certified margin had a violation rate below its target on unseen objectives. Moreover, at equal cost, SRC’s correction had fewer or equal infeasible solutions to the baselines in 57 of 62 comparisons. The SRC framework applies to any setting in which a surrogate model chooses a decision whose feasibility can be verified for a sample of instances.
\end{abstract}


\section{Introduction}\label{sec1}

Many real-world optimization problems involve constraints that do not have known closed-form expressions. This setting commonly arises across domains such as radiation therapy planning, humanitarian logistics, energy dispatch, virtual power plants, product pricing, and routing and scheduling \cite{bib1, bib2, bib3, bib4, bib5}. The standard approach in such a setting is to learn a surrogate model from the available data, and optimize against the surrogate instead \cite{bib6, bib7}. This paradigm --- embedding a learned model as a constraint inside an optimization problem --- is known as \textit{Optimization with Constraint Learning (OCL)} \cite{bib8}.

However, a learned surrogate can differ from the true constraint in the regions crucial for the optimization problem. Since the optimizer converges over the surrogate rather than the true function, it can return a solution that satisfies the surrogate while violating the true constraint. At the time of deployment, there is usually no reliable way to verify this since the true constraint remains unknown. Optimizing against a surrogate therefore offers no guarantee that the returned decision is actually feasible. Hence, in such a setting, the focus should be to ensure the true feasibility of the  decision the optimizer returns, rather than improving the average accuracy of the surrogate \cite{bib9}.

For example, in our experiments, we see scenarios where a pipe design accepted by the surrogate can leave a junction of a water network below its minimum pressure, and an operating point accepted by the surrogate can fall short of the required production of a reactor. Such surrogates are usually trained once and then reused for many objectives with varying prices, costs or demands. However, checking each decision with the simulator online is often too slow or too costly. Hence, our setting relies on checking a sample of decisions offline, before the surrogate is deployed.

Conformal prediction \cite{bib10, bib11} computes a safety margin for a model's raw predictions using held-out data and provides a distribution-free coverage guarantee. Within OCL, this idea is adapted by embedding conformally calibrated constraints directly into the optimization problem, so that the resulting feasible region inherits a statistical feasibility guarantee \cite{bib12}. Like any conformal guarantee, however, this relies on \textit{exchangeability} between the test point and the calibration data: coverage holds only when the test point can be treated as drawn at random from the same distribution as the data used to compute the calibration margin \cite{bib10}.

However, we show that the point returned by an optimization solver does not satisfy this condition. It is not a random point in the data, but is instead the  cheapest feasible point satisfying the constraint,
\[
  x^{\star}(c)\in\operatorname*{arg\,min}_{x}\; c(x)
  \quad\text{subject to}\quad \hat{h}(x)+\hat{q}\le 0,
\]
where $\hat{h}$ is the learned surrogate, $\hat{q}$ is the conformal margin, and $c$ is the objective. 

The returned decision $x^{\star}(c)$ depends upon the objective and is influenced by the boundary of the feasible region, built from the margin $\hat q$. It therefore differs in distribution from the calibration data, which is drawn at random (Section~\ref{sec:fcs}). Through our experiments, we show that this has major effects on the coverage guarantee: a margin that covers about $90\%$ of random inputs covers as few as $17\%$ of the decisions the optimizer returns (Section~\ref{sec:exp-fcs}). The dependence on $\hat q$ is a case of \textit{feedback covariate shift}, where a model's own predictions determine which inputs are subsequently evaluated \cite{bib13}. Even though feedback covariate shift is the closest existing problem family to our setting, its standard reweighting solution does not apply to our problem (Section~\ref{sec:fcs}, Proposition~\ref{thm:t2}).

We propose \emph{selected-risk calibration} (SRC), a framework where we calibrate the learned constraint against the decisions the optimizer returns, rather than against random points. As the name suggests, its unit of calibration is the \emph{selected risk}, the probability that the decision returned for a random objective violates the true constraint. Its guarantee is finite-sample and distribution-free (Theorem~\ref{thm:t4}). In our experiments, none of the certified margins we checked on new objectives exceeded its target risk, and at equal cost SRC's correction had fewer or equal violations than the baseline in 57 of 62 comparisons (Section~\ref{sec:exp}).

Although we introduce SRC as a calibration framework for optimization with constraint learning, its coverage does not depend on anything specific to this setting. It applies whenever a fixed procedure depends on a learned model to reach a final output, and whether that output's feasibility can be checked for a sample of instances.

We make the following contributions:
\begin{itemize}[noitemsep, topsep=4pt, partopsep=0pt, parsep=2pt]
  \item We prove that random-point conformal coverage does not, by itself, bound the probability of infeasibility at the optimizer's returned decision (Theorem~\ref{thm:separation}).
  \item We prove that when the residuals (the difference between the true constraint and the learned surrogate) have a continuous distribution, the distribution of the optimizer's returned decision is \emph{mutually singular} with the distribution of the calibration data for almost every calibration sample whenever the learned constraint is active at the returned decision and the correction does not depend on the calibration sample. Hence, the likelihood-ratio reweighting method that restores conformal validity under feedback covariate shift cannot be used here (Proposition~\ref{thm:t2}).
  \item We propose the selected-risk calibration (SRC) framework, which calibrates the constraint against the optimizer's returned decisions, and prove that, with probability of at least $1-\delta$, every margin it certifies has selected risk at most its target (Theorem~\ref{thm:t4}).
  \item Our experiments show that conformal coverage is largely lost at the returned decisions (Section~\ref{sec:exp-fcs}) and that the SRC certified margins stayed within their target risks on unseen objectives (Section~\ref{sec:exp-cert}). We then evaluate SRC's correction on three simulation benchmarks at matched cost and show that it had a lower or equal violation rate than the baselines in 57 of 62 comparisons (Section~\ref{sec:exp-cost}).
\end{itemize}

\section{Related Work}\label{sec2}

\textbf{Optimization with constraint learning.} OCL fits a machine-learning model to constraint data and embeds it in the optimization problem \cite{bib12, bib8}. This paradigm is prevalent in settings where the relationship between decision and output variables is either unknown or in surrogate modeling scenarios where it is known but difficult to optimize \cite{bib15, bib16}. A wide variety of surrogate models have been used for this task, including piecewise-linear neural networks \cite{bib6}, gradient boosting \cite{bib7}, and tree ensembles \cite{bib17}, or any other regressor embeddable in the optimization problem. In such a setting, it is crucial to pay attention to the feasibility of the final decision and to how the learned model influences it. Commonly employed methods like trust regions keep the optimizer's search close to the training data \cite{bib22, bib21} and ensembles average several surrogates, as in the W-MICL variant of \cite{bib12}, but neither gives a statistical guarantee on the returned decision.

Smart ``Predict, then Optimize'' (SPO) \cite{bib9} and decision-focused learning (DFL) \cite{bib18} study the same paradigm where an ML model's prediction is a component of the optimization problem. They focus on training the ML model in an end to end setting where the decision error induced by the prediction is minimized. On the other hand, SRC calibrates the model on the solver's returned decision after it has been trained regularly, giving a finite-sample bound on how often the resulting decisions are infeasible. An ML model trained end to end to minimize the induced optimization error can still be calibrated using SRC.

\textbf{Conformal prediction} uses held-out data to give a finite-sample, distribution-free coverage guarantee for subsequent predictions \cite{bib10, bib11}. The most direct adaptation of conformal prediction to OCL is the C-MICL framework \cite{bib12}, which embeds a conformal margin directly into a mixed-integer optimization problem and proves that the resulting solution is feasible with probability at least $1-\alpha$, under the assumption that, given the true feasibility of a point, its feasibility in the optimization problem and its conformal coverage are conditionally independent. Some related works embed conformal prediction as chance constraints or uncertainty sets rather than deterministic margins \cite{bib23, bib24, bib25}. However, in all these works, the object receiving a coverage guarantee is a random point drawn independently of the optimization procedure, and hence, doesn't give a guarantee for the decision the optimizer returns.

\textbf{When exchangeability fails.} Several frameworks exist for providing guarantees when the test point is not exchangeable with the calibration data. Tibshirani et al.\ \cite{bib26} compute weights for conformal prediction from the likelihood ratio between the test and calibration covariate distributions. Barber et al.\ \cite{bib14} use weighted quantiles to introduce robustness against distribution shift. Closer to our setting, Fannjiang et al.\ \cite{bib13} study feedback covariate shift, where a machine learning model's predictions influence the downstream test data, similar to the surrogate constraint's role in an optimization problem, and show that validity can be restored by reweighting calibration scores by the likelihood ratio between the selected and the calibration inputs. Their guarantee requires the distribution of the selected input to be absolutely continuous with respect to the calibration distribution. However, OCL differs from this setting in critical ways. When a learned constraint is active while the solver converges in the feasible space, the returned decision is a single point at the boundary of this space. Hence, as explained in Proposition~\ref{thm:t2}, the distribution of an optimizer's returned decision has no density with respect to the input distribution, and its standard reweighting solution cannot be implemented here.

\section{Problem Setup}
 
\subsection{Optimization with a learned constraint}\label{sec:ocl}
 
Let $\mathcal{X}\subset\mathbb{R}^{d}$ be a compact decision space and
let $h:\mathcal{X}\to\mathbb{R}$ be a constraint such that a decision $x$ is \emph{feasible} if $h(x)\le 0$. We are interested in the optimization problems
\begin{equation}\label{eq:true-problem}
  \min_{x\in\mathcal{X}}\ c(x)
  \qquad\text{subject to}\qquad
  h(x)\le 0,
\end{equation}
where $h$ has no known closed form equation and can only be evaluated through an expensive oracle, such as a simulator or a physical experiment. Optimization with constraint learning (OCL) replaces $h$ by a continuous surrogate $\hat{h}$ fitted to oracle data
\cite{bib8,bib22} and solves
\begin{equation}\label{eq:surrogate-problem}
  \min_{x\in\mathcal{X}}\ c(x)
  \qquad\text{subject to}\qquad
  \hat{h}(x)\le 0.
\end{equation}
 
We call $s=h-\hat{h}$ the \emph{residual} of the surrogate. It is \emph{pessimistic} when $s(x)<0$ and \emph{optimistic} when $s(x)>0$. When the residual is pessimistic, any point the surrogate accepts is feasible, though possibly more conservative than necessary. On the other hand, when it is optimistic, the surrogate is more lenient than the true constraint. Hence, any decision with $\hat{h}(x)\le 0<h(x)$ is accepted by \eqref{eq:surrogate-problem} but is infeasible in reality.
 
\subsection{The conformal margin}\label{sec:margin}
 
Let $\mathcal{D}_{\mathrm{cal}}=(X_{1},\dots,X_{m})$ be a calibration sample drawn i.i.d.\ from a distribution $P$ on $\mathcal{X}$,
independently of the data used to fit $\hat{h}$, and let
$S_{i}=s(X_{i})$. Given a level $\alpha\in(0,1)$ with $\alpha>1/(m+1)$,
split conformal prediction \cite{bib10, bib11} sets
\begin{equation}\label{eq:margin}
  k=\bigl\lceil(1-\alpha)(m+1)\bigr\rceil ,
  \qquad
  \hat{q}=S_{(k)} ,
\end{equation}
where $S_{(k)}$ is the $k$-th smallest of $S_{1},\dots,S_{m}$. The
condition on $\alpha$ ensures $k\le m$.
If $X_{m+1}\sim P$ is a further draw, independent of
$\mathcal{D}_{\mathrm{cal}}$, then
\begin{equation}\label{eq:coverage}
  \Pr\bigl(h(X_{m+1})\le\hat{h}(X_{m+1})+\hat{q}\bigr)\;\ge\;1-\alpha ,
\end{equation}

where the probability is over $\mathcal{D}_{\mathrm{cal}}$ and $X_{m+1}$. This is marginal coverage averaged over calibration samples. The argument requires only that $S_{1},\dots,S_{m+1}$ be exchangeable, and when these scores are almost surely distinct the left-hand side equals $k/(m+1)$. Consequently, a random point $X_{m+1}$ is accepted by the deployed constraint yet infeasible with probability at most $\alpha$. An accepted point that is infeasible satisfies $h(X_{m+1})>\hat{h}(X_{m+1})+\hat{q}$.
 
\subsection{The deployed set and the returned decision}\label{sec:deployed}
 
Later sections add a continuous \emph{correction} $g:\mathcal{X}\to\mathbb{R}$ to the margin, so we define the
\emph{deployed set} in that generality:
\begin{equation}\label{eq:deployed-set}
  \widehat{\mathcal{X}}(g)
  =\bigl\{x\in\mathcal{X}:\hat{h}(x)+\hat{q}+g(x)\le 0\bigr\}.
\end{equation}

The choice $g\equiv 0$ is the plain conformal deployment. Since $\hat{h}$
and $g$ are continuous, $\widehat{\mathcal{X}}(g)$ is compact.
 
A surrogate and its margin are usually fitted once and then used to solve many optimization problems that might have different objective functions. We therefore treat the objective as random, drawing $c$ from a distribution $Q$ over a family $\mathcal{C}$ of continuous functions on $\mathcal{X}$. The distribution $Q$ may depend on the data used to build the deployed set, but given that data, $c$ is drawn independently of everything else. The solver returns
\begin{equation}\label{eq:returned}
  x^{\star}(c)\in\operatorname*{arg\,min}_{x\in\widehat{\mathcal{X}}(g)} c(x).
\end{equation}
A minimizer exists when $\widehat{\mathcal{X}}(g)$ is non-empty. If there are several, $x^\star(c)$ is any one of them, chosen measurably in $c$. This exact minimizer is used for the theory, while the certificate applies to whatever decision the solver returns.
 
The \emph{selected risk} of the correction $g$ is the probability that
this decision is infeasible,
\begin{equation}\label{eq:selected-risk}
  R(g)=\Pr_{c\sim Q}\bigl(h(x^{\star}(c))>0\bigr),
\end{equation}
with $R(g)=1$ when $\widehat{\mathcal{X}}(g)$ is empty. For a tolerance $\tau\ge 0$, we write $R_\tau(g)$ for the same probability with $h(x^\star(c))>\tau$. The probability is over $c$ alone, with the calibration sample held fixed, so $R(g)$ depends on $\mathcal{D}_{\mathrm{cal}}$ through $\hat{q}$, and through $Q$ if $Q$ also depends on it.
 
The probabilities in \eqref{eq:coverage} and \eqref{eq:selected-risk} concern different points. Coverage is about $X_{m+1}$, a fresh draw from $P$. Selected risk is about $x^{\star}(c)$, which is not drawn from $P$ at all: it minimizes a cost over a feasible set built from $\hat{q}$. The rest of the paper is
about the gap between the two.

\section{Coverage Guarantee at the Returned Decision}
\label{sec:fcs}
 
\subsection{Coverage does not constrain the selected risk}\label{sec:sep}
 
To see why the two probabilities can differ, recall how $\hat{q}$ is chosen. It is a single number, set so that the residual exceeds it at roughly an $\alpha$ fraction of points drawn from $P$, and it carries no information about where in $\mathcal{X}$ those points lie. The solver, on
the other hand, does not sample from $P$. When the learned constraint is active at the optimum, the solver returns a point on the boundary of the deployed set influenced by the objective. If the residual is large at the points the objective favors, the constraint along with its margin will be violated there regardless of how well $\hat{q}$ is
calibrated. The following theorem states this precise with an instance in which coverage holds exactly, while the selected risk can take any value in $[0,1]$.
 
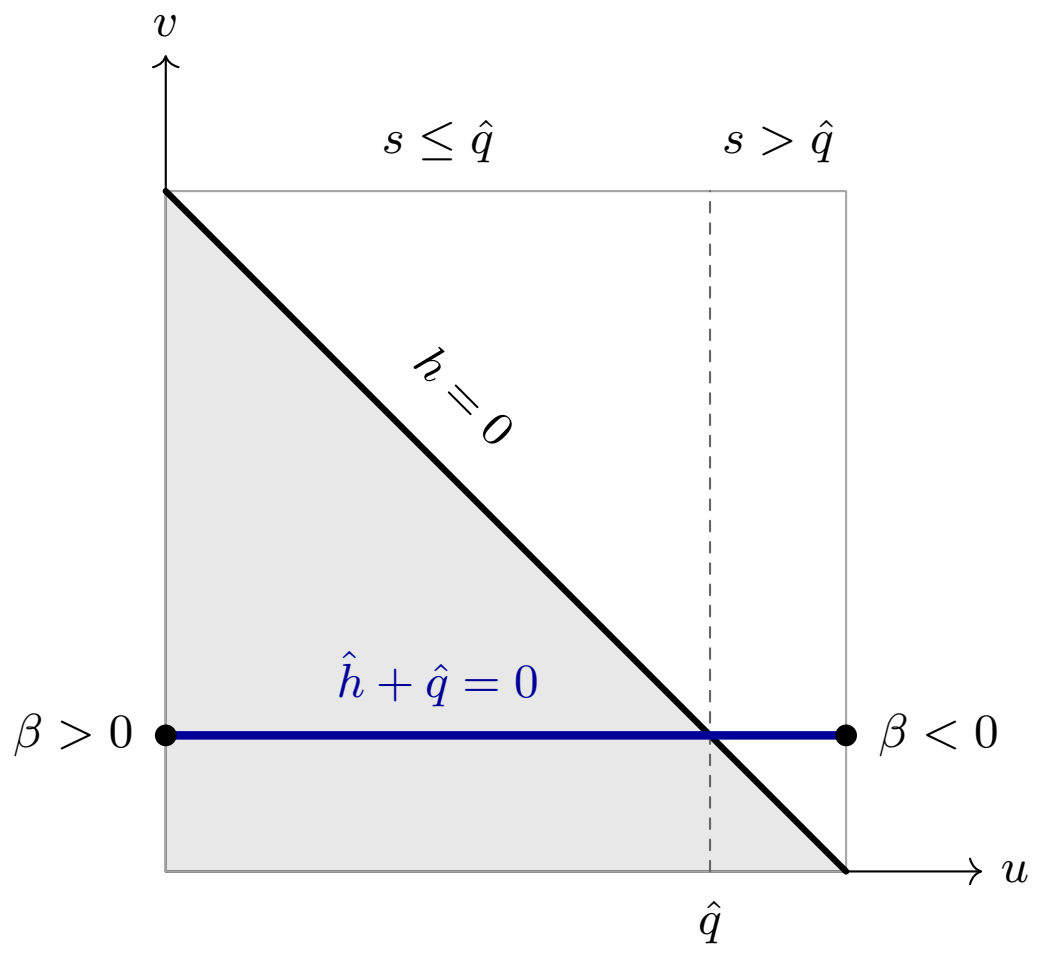
\begin{figure}[t]
\centering
\begin{tikzpicture}[scale=4.5, line cap=round, line join=round]
  \fill[black!9] (0,0) -- (1,0) -- (0,1) -- cycle;
  \draw[->, thin] (0,0) -- (1.2,0) node[right] {\small $u$};
  \draw[->, thin] (0,0) -- (0,1.2) node[above] {\small $v$};
  \draw[thin, black!35] (0,0) -- (1,0) -- (1,1) -- (0,1) -- cycle;
  \draw[very thick] (1,0) -- (0,1);
  \node[rotate=-45, font=\small] at (0.44,0.70) {$h=0$};
  \draw[thin, black!60, dash pattern=on 2pt off 2pt] (0.8,0) -- (0.8,1);
  \node[anchor=north, font=\small] at (0.8,-0.015) {$\hat{q}$};
  \node[anchor=south, font=\small] at (0.40,1.015) {$s\le\hat{q}$};
  \node[anchor=south, font=\small] at (0.90,1.015) {$s>\hat{q}$};
  \draw[line width=1.6pt, blue!60!black] (0,0.2) -- (1,0.2);
  \node[anchor=south, font=\small, blue!60!black] at (0.40,0.215)
       {$\hat{h}+\hat{q}=0$};
  \fill (0,0.2) circle (0.016);
  \node[anchor=east, font=\small] at (-0.02,0.2) {$\beta>0$};
  \fill (1,0.2) circle (0.016);
  \node[anchor=west, font=\small] at (1.02,0.2) {$\beta<0$};
\end{tikzpicture}
\caption{The construction behind Theorem~\ref{thm:separation}, drawn
with $\hat{q}=0.8$. The shaded triangle is the true feasible region
$\{h\le 0\}$, and the horizontal line $\hat{h}+\hat{q}=0$ is the upper
edge of the deployed set $\{\hat{h}+\hat{q}\le 0\}$. The vertical line
$u=\hat{q}$ separates the points where the conformally padded surrogate
bounds $h$ from those where it does not; the strip to its right has
probability $1-\hat{q}$ under $P$. Every objective drives the solver onto
the horizontal line, and the sign of $\beta$ decides which end.
Objectives with $\beta<0$ all land in the uncovered strip.}
\label{fig:instance}
\end{figure}
 
\begin{theorem}[Exact coverage does not constrain the selected risk]
\label{thm:separation}
Let $\alpha\in(0,1)$ and $p\in[0,1]$. There exist a compact set
$\mathcal{X}\subset\mathbb{R}^{2}$, a distribution $P$ on $\mathcal{X}$,
affine functions $h$ and $\hat{h}$, and a distribution $Q$ over linear
objectives, none of which depend on $\alpha$, such that for every
calibration size $m$ with $\alpha>1/(m+1)$, the plain conformal deployment
($g\equiv 0$) has the following two properties.
\begin{enumerate}
  \item\label{it:cov} The residuals $S_{1},\dots,S_{m+1}$ are i.i.d.\
  uniform on $[0,1]$, and the probability in \eqref{eq:coverage} equals
  $k/(m+1)$. This is $1-\alpha$ when $(1-\alpha)(m+1)$ is an
  integer, or lies in $\bigl(1-\alpha,\,1-\alpha+\tfrac{1}{m+1}\bigr)$
  otherwise.
  \item\label{it:risk} With probability one over
  $\mathcal{D}_{\mathrm{cal}}$, the deployed set is non-empty, the
  minimizer in \eqref{eq:returned} is unique for $Q$-almost every
  objective, and $R(0)=p$.
\end{enumerate}
\end{theorem}
 
\emph{Proof sketch.} The construction is shown in
Figure~\ref{fig:instance}. The decision set is the unit square with $P$ uniform, the true constraint is $h(u,v)=u+v-1$, and the surrogate is $\hat{h}(u,v)=v-1$. The residual is then $s(u,v)=u$, which is uniform on $[0,1]$ under $P$ and gives the coverage in part~\ref{it:cov} of the theorem. The objectives are $c_{\beta}(u,v)=\beta u-v$. Every objective prefers larger values of $v$, so the solver always returns a point on the top edge of the deployed set, $v=1-\hat{q}$ , and the sign of $\beta$ decides whether that point is the left corner $u=0$ or the right corner $u=1$. Along this edge the true constraint equals $h=u-\hat{q}$, which holds true at the left corner and, because $\hat{q}<1$ almost surely, is violated at the right one. Taking $Q$
with $Q(\beta<0)=p$ therefore gives $R(0)=p$. Note that nothing in the construction involves $\alpha$. Hence, raising the coverage level to make the margin safer does not lower the selected risk. This tells us that coverage at random inputs and the probability of the returned decision being feasible are different quantities.  
 
Appendix~\ref{app:t1} gives the full proof, shows that the conclusion does not depend on the surrogate being optimistic everywhere, and discusses what the theorem implies for bounds on the selected risk. Section~\ref{sec:exp-fcs} empirically shows the same effect.
 
\subsection{Likelihood-ratio reweighting does not apply}\label{sec:reweight}
 
As we saw, the returned decision is not exchangeable with the calibration sample because it is chosen using the margin that the sample produced. This is an instance of feedback covariate shift \cite{bib13}, in which the data
determine the distribution of the test input. In that setting, Fannjiang et al.\ \cite{bib13} restore conformal coverage by reweighting the calibration scores with the likelihood ratio between the test-input distribution and
$P$. However, their guarantee assumes that the test-input distribution is absolutely continuous with respect to $P$. Here the test input is the returned decision, and over random objectives, its distribution is
\[
  \tilde P(A)=\Pr_{c\sim Q}\bigl(x^{\star}(c)\in A\bigr),
  \qquad A\subseteq\mathcal{X}.
\]

When the learned constraint is active at the returned decision, this distribution is concentrated on the boundary of the deployed set, and we now show that that boundary has probability zero under $P$ for almost every calibration sample (Proposition~\ref{thm:t2}).
 
\begin{proposition}[Likelihood-ratio reweighting is unavailable at the returned decision]
\label{thm:t2}
Let the correction $g$ be fixed independently of
$\mathcal{D}_{\mathrm{cal}}$, and let the residual $s(X)$, $X\sim P$,
have a continuous distribution. Suppose the learned constraint is active
at the returned decision:
\[
  \hat{h}\bigl(x^{\star}(c)\bigr)+\hat{q}+g\bigl(x^{\star}(c)\bigr)=0
  \qquad\text{for $Q$-almost every $c$.}
\]
Then, for almost every calibration sample with a non-empty deployed set, $\tilde P$ and $P$ are mutually singular. The set
\[
  M=\bigl\{x\in\mathcal{X}:\hat{h}(x)+\hat{q}+g(x)=0\bigr\}
\]
satisfies $\tilde P(M)=1$ and $P(M)=0$. In particular $\tilde P$ is not
absolutely continuous with respect to $P$, so the likelihood ratio used
by \cite{bib13} does not exist and their coverage guarantee does not
apply.
\end{proposition}

\emph{Proof sketch.} The equality $\tilde P(M)=1$ is simply the active-constraint assumption that every returned decision lies on $M$. For $P(M)=0$,
note that the level sets $\{\hat{h}+g=-t\}$, $t\in\mathbb{R}$, are disjoint, so $P$ gives positive probability to at most countably many of them. Since the residuals have a continuous distribution, it avoids those countably many values with probability one. The full proof is in Appendix~\ref{app:t2}, together with
conditions under which the conclusion holds for every calibration sample.

The conclusion does not need every returned decision to be active. If the constraint is only active for a set of objectives of probability $a>0$, the same argument gives $\tilde P(M)\ge a>0=P(M)$, so $\tilde P$ is still not absolutely continuous with
respect to $P$. For example, in the certification study of Section~\ref{sec:exp-cert}, the deployed constraint is active, to within $10^{-6}$, at 64\% of the returned decisions on methane and 70\% on Williams--Otto.
 
In the instance of Theorem~\ref{thm:separation}, the returned decision is always one of the corners $(0,1-\hat{q})$ and $(1,1-\hat{q})$, so $\tilde P$ puts all of its mass on two points, irrespective of however smoothly $Q$ is
spread over objectives. The singularity comes from the optimization process itself, not from a particular choice of $Q$.
 
Proposition~\ref{thm:t2} does not say that covariate-shift corrections are wrong, only that the returned decision lies outside the scope of likelihood-ratio reweighting. Reweighting calibration scores by a likelihood ratio cannot certify an optimizer's returned decisions because $\tilde P$ is not a reweighting of $P$. This proposition applies whenever the learned constraint is active at the returned decision and the correction does not depend on the calibration data. A guarantee for the returned decision therefore has to be built around the decision itself, which is the approach we take in Section~\ref{sec:method}.

\section{Methodology}
\label{sec:method}
 
\subsection{Calibrating the returned decision}
 
Selected-risk calibration certifies the selected risk $R(g)$ in
\eqref{eq:selected-risk}. Therefore, its unit is a random objective together with the decision the solver returns for it.
 
From here on, $x^{\star}(c)$ denotes the point our solver actually returns, which might not necessarily be the global minimizer. For our experiments, a solver that finds no feasible point counts as a violation. Given its starting points our solver is deterministic, and its starting points are drawn for each objective from that objective's own seed. Hence, different objectives give independent trials, and we treat the starting points as part of the random objective. With these conventions, $R(g)$ is the probability that the deployed pipeline, solver included, either returns an infeasible decision or none at all.

\paragraph{Candidate margins.}

Our candidates replace the constant margin $\hat{q}$ by
\begin{equation}\label{eq:cand-margin}
  m_{\theta,\eta}(x)
  =\eta\Bigl((1-\theta)+\theta\,\frac{\hat{\sigma}(x)}{\bar{\sigma}}\Bigr),
\end{equation}
where $\hat{\sigma}(x)>0$ is the local-uncertainty estimate of C-MICL~\cite{bib12}, a network trained on the training data to predict the size of the residual $|h-\hat h|$ and floored at a small positive
constant. $\bar{\sigma}$ is its median over the calibration inputs, the shape $\theta\in[0,1]$ decides how closely the margin follows $\hat{\sigma}$, and the size $\eta>0$ sets its overall level. In the notation of Section~\ref{sec:deployed}, deploying
$\hat{h}(x)+m_{\theta,\eta}(x)\le 0$ means using the correction $g=m_{\theta,\eta}-\hat{q}$. The shape $\theta=0$ gives a constant margin, which for $\eta=\hat{q}$ is the plain conformal deployment when $\hat q>0$, and $\theta=1$ gives a margin proportional to C-MICL's uncertainty estimate.

 The guarantee does not depend on this particular family, and holds for any family fixed before the certification objectives are drawn. For example, Section~\ref{sec:exp-cost} uses a correction learned directly from the errors at returned decisions. It adds a multiple of a smooth function, fitted to lie above of the remaining conformal residuals $h-\hat h-\hat q$, to $\hat h+\hat q$ (Appendix~\ref{app:matched}). Any such family can take the place of $m_{\theta,\eta}$ in the following procedure.
 
\paragraph{The procedure.}

Candidates are compared at equal cost, and each step uses its own set of
objectives drawn from $Q$. For a given target cost, the size of each shape
$\theta$ in $\Theta$ is tuned on $\mathcal{D}_{\mathrm{match}}$ until the average cost of the returned decisions is within a set tolerance of the target ($1\%$ in Section~\ref{sec:exp-cert}). Shapes that cannot be matched are dropped. The shape with the lowest violation rate on $\mathcal{D}_{\mathrm{prop}}$ is chosen along with the resulting margin. Following that, the $n$ objectives of $\mathcal{D}_{\mathrm{cert}}$ are solved. If $N$ of them end in a violation, we compute the one-sided Clopper--Pearson upper bound~\cite{bib32}
\begin{equation}\label{eq:cp}
  \bar{U}=U_{n}(N;\varepsilon),\qquad
  U_{n}(v;\varepsilon)
  =\sup\bigl\{r\in[0,1]:\Pr\bigl(\mathrm{Bin}(n,r)\le v\bigr)\ge\varepsilon\bigr\},
\end{equation}
and call the margin \emph{certified} at the risk target $\gamma$ if
$\bar{U}\le\gamma$. For $v<n$, $U_{n}(v;\varepsilon)$ is the
$(1-\varepsilon)$-quantile of the $\mathrm{Beta}(v+1,n-v)$ distribution, and
$U_{n}(n;\varepsilon)=1$. A fourth set of unseen objectives is used to check the certificates in Section~\ref{sec:exp-cert}. Algorithm~\ref{alg:src} summarizes the procedure.
 
\begin{algorithm}[!ht]
\caption{Selected-Risk Calibration (SRC) at one target cost}
\label{alg:src}
\begin{algorithmic}[1]
\Require surrogate $\hat{h}$, uncertainty estimate $\hat{\sigma}$ and its calibration median $\bar{\sigma}$, shape grid $\Theta$, target cost $C$ with cost tolerance $\tau_C$, disjoint objective sets $\mathcal{D}_{\mathrm{match}}$, $\mathcal{D}_{\mathrm{prop}}$, $\mathcal{D}_{\mathrm{cert}}$ drawn from $Q$, with $|D_{\mathrm{cert}}|=n$ and $n/L$ objectives from each family, risk target $\gamma$, violation tolerance $\tau$, and level $\varepsilon$
\Ensure a locked margin $m$, its statistic $\bar U$, and whether it is certified or \textsc{no match}
\For{each shape $\theta\in\Theta$}
  \State choose $\eta_{\theta}$ so that the mean cost on
    $\mathcal{D}_{\mathrm{match}}$ under $m_{\theta,\eta_{\theta}}$ is within $\tau_C$ of $C$. If no such $\eta_\theta$ is found, drop $\theta$.
\EndFor
\If{no shape remains} \Return \textsc{no match} \EndIf
\State $\hat{\theta}\gets$ the remaining shape with the fewest violations on $\mathcal{D}_{\mathrm{prop}}$
  \Comment{ties go to the smaller $\theta$}
\State lock $m\gets m_{\hat{\theta},\eta_{\hat{\theta}}}$
  \Comment{$\mathcal{D}_{\mathrm{cert}}$ is first used after this line}
\State $N\gets$ number of $c\in\mathcal{D}_{\mathrm{cert}}$ whose solve under
  $m$ fails or returns $x^{\star}(c)$ with $h(x^\star(c))>\tau$
\State $\bar{U}\gets U_{n}(N;\varepsilon)$
  \Comment{Eq.~\eqref{eq:cp}}
\State \Return $m$ and $\bar{U}$; $m$ is certified if $\bar{U}\le\gamma$
\end{algorithmic}
Across target costs, SRC returns the cheapest certified margin or abstains if none is certified.
\end{algorithm}

\subsection{Finite-sample guarantee}
 
Our study certifies a margin for each target cost in each run. We index them by $k=1,\dots,K$ and write $g_{1},\dots,g_{K}$ for the corresponding corrections. Margin $k$ has $N_{k}$ violations among its $n$ certification objectives and bound $\bar{U}_{k}=U_{n}(N_{k};\varepsilon)$,
and is certified if $\bar{U}_{k}\le\gamma$. Margins from the same run share their certification objectives. In our experiments the objectives come from $L=5$ families, and every certification set holds the same number, $n/L$. We therefore take $Q=\frac{1}{L}\sum_{\ell=1}^{L}Q_{\ell}$ to be the equal mixture of the families' distributions $Q_{1},\dots,Q_{L}$. The $Q_{\ell}$ may depend on the
data used to build the candidates.
 
\begin{theorem}[Validity of selected-risk certification]
\label{thm:t4}
For each margin $k$, let $g_{k}$ be built from data that exclude the margin's certification objectives, and suppose that, given this data (which also
fix $Q_{1},\dots,Q_{L}$), the $n$ certification objectives are independent,
with $n/L$ of them drawn from each $Q_{\ell}$. Let $b$ be the largest
$v\in\{0,\dots,n\}$ with $U_{n}(v;\varepsilon)\le\gamma$, or $b=-1$ if
there is none, and suppose that $b\le n\gamma-1$. Then, for every margin,
\[
  \Pr\bigl(\text{margin $k$ is certified and } R_\tau(g_k)>\gamma\bigr)
  \le\varepsilon .
\]
Consequently, if $K\varepsilon\le\delta$, the probability that some margin is
certified while its selected risk exceeds $\gamma$ is at most $\delta$.
\end{theorem}

In our experiments we use $n=300$, $\gamma=0.12$ and $\varepsilon=0.05/40$, where $40$ is the number of
margins in the study (Section~\ref{sec:exp-cert}), so $K\varepsilon\le 0.05$. A margin is then certified exactly when it has at most $b=19$ violations, comfortably below $n\gamma-1=35$. Note that Theorem~\ref{thm:t4} uses $U_n(N;\varepsilon)$ only as a test of the fixed target $\gamma$. This test remains valid when the families differ in risk, even though $U_n(N;\varepsilon)$ need not then be an upper confidence bound for the selected risk.

\emph{Proof sketch.} Given the data that built $g_{k}$, each certification
objective is an independent trial whose probability of ending in a
violation depends on its family, and these probabilities average to $R_\tau(g_k)$. Since $U_{n}(v;\varepsilon)$ increases with $v$, the margin is
certified when its number of violations is at most $b$. If $R_\tau(g_k)>\gamma$, then $b\le nR_\tau(g_k)-1$, and a classical comparison due to
Hoeffding~\cite{bib33} shows that the number of violations, a sum of
independent trials with mean $nR_\tau(g_k)$, is not more likely to fall to $b$ or below a
binomial count with the same mean. That binomial
probability is below $\varepsilon$ by the definition of $U_{n}$, because
$R_\tau(g_k)>\gamma\ge U_{n}(b;\varepsilon)$. A union bound over the margins
gives the last claim. The full proof is in Appendix~\ref{app:t4}.
 
As mentioned before, the guarantee does not rely on which shape is selected, so the shape of the margin is a design choice, which we leave to held-out objectives. A constant margin needs to be large enough to protect the worst decision the solver could return, hurting the optimality of every objective. On the other hand, a margin that grows only where the surrogate is less accurate can give up less objective value at the returned decisions. Whether this pays off depends on how unevenly the surrogate errs, which Sections~\ref{sec:exp-cert} and~\ref{sec:exp-cost} examine empirically.

\section{Experiments}
\label{sec:exp}

The experiments focus on three questions. Does a split-conformal margin, that covers random inputs at its nominal rate, cover the decisions an
optimizer returns (Section~\ref{sec:exp-fcs})? When SRC certifies a margin at risk target $\gamma$, is the margin's selected risk on new objectives at
most $\gamma$ (Section~\ref{sec:exp-cert})? How does the correction SRC uses compare with existing methods at equal deployment cost (Section~\ref{sec:exp-cost})? 

\subsection{Setup}
\label{sec:exp-setup}

We use three benchmarks, each having one learned constraint $h(x)\le 0$. The first is the methane membrane reactor of C-MICL \cite{bib12}, where the constraint requires a minimum benzene output, computed by integrating the reactor model. The second, Williams--Otto plant \cite{bib27}, requires a minimum product flow, computed by solving the plant's recycle loop. The third is the Hanoi water network \cite{bib28, bib29},  which requires a minimum pressure at every junction, computed by a hydraulic simulation. The first two benchmarks have five decision variables and the third has eight. A more extensive description of the benchmarks is given in Table ~\ref{tab:benchmarks}. Apart from Section~\ref{sec:exp-fcs}, every objective used in the study is linear, $c(x)=w^{\top}x$, with $w$ drawn from one of five families. Every set of objectives holds the same number from each family. Lastly, a returned decision $x^{\star}$ violates the constraint when $h(x^{\star})>0$, and we estimate the selected risk of a margin by the fraction of objectives whose returned decision violates the constraint. Appendix~\ref{app:details} describes the variables, the objective families, the surrogates, the solver and all data sizes in detail.

\subsection{Split conformal does not protect the returned decision}
\label{sec:exp-fcs}

We first deploy the surrogate with the plain conformal margin, $\hat h+\hat q\le 0$ (Section~\ref{sec:margin}), on all three benchmarks,
with five seeds and 200 objectives per seed. Without any margin, the returned decision violates the true constraint for a fraction $0.488$ of the objectives on methane, $0.366$ on Williams--Otto and $0.507$ on Hanoi, averaged over seeds. With the conformal margin, these fractions are $0.485$, $0.352$ and $0.485$, almost unchanged. Yet on the same fits the margin covers between $0.889$ and $0.940$ of 800 random held-out inputs, close to its nominal $0.90$. Overall, the margin holds for random inputs, and has very little effect over the decisions the optimizer returns. Moreover, in 6 out of 15 runs, the margin gives more violations than the constraint without the margin (Table~\ref{tab:nomargin} in Appendix~\ref{app:fcs} gives the entire run).


We ran a study on the Methane and Williams--Otto benchmarks that measures the coverage of the same margin $m$ on three sets of points: random held-out inputs, the decisions returned for 100 random objectives, and \textit{matched-boundary points}, which are random inputs that the optimizer did not choose but that lie as close to the boundary as the returned decisions. A point $x$ is covered when $h(x)\le\hat h(x)+m(x)$. If coverage is lost at the returned decisions and not at the matched-boundary points, this points to the optimizer's choice rather than to a margin that is weak everywhere near the boundary. We run six configurations: two surrogates, a
symbolic expression and a neural network (MLP), combined with two margins, the scalar margin $m(x)=\hat q$, which is constant in $x$, and the normalized margin of C-MICL, which scales with a learned estimate of the surrogate's error. Methane uses all four combinations and Williams--Otto the scalar margin only. Each configuration is repeated five times at $\alpha=0.10$
(Appendix~\ref{app:fcs}).

\begin{figure}[t]
  \centering
  \includegraphics[width=\textwidth]{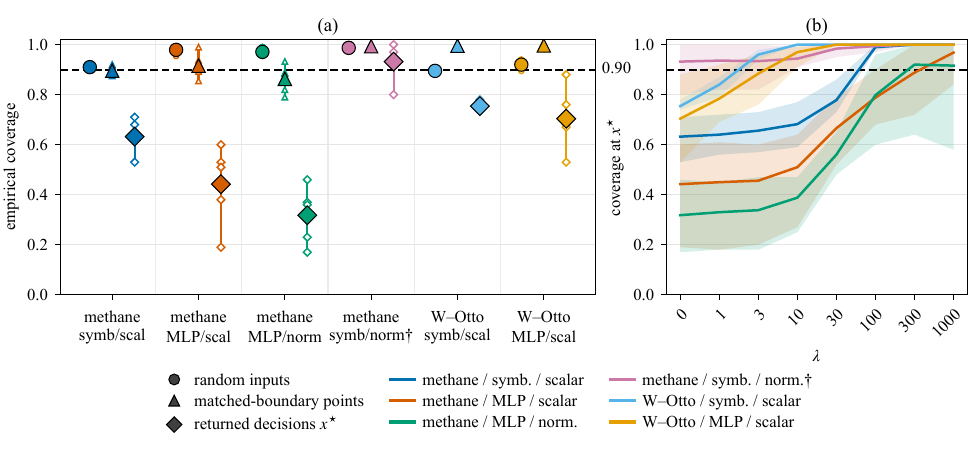}
  \caption{Coverage of a split-conformal margin at level $0.90$, five repetitions per configuration. \textbf{(a)} Large markers are means and small markers represent each single run ($\dagger$: see Appendix~\ref{app:fcs}). \textbf{(b)}~Coverage
  at the returned decision as the penalty driving the optimizer away from the boundary grows with
  $\lambda$.}
  \label{fig:fcs-collapse}
\end{figure}

The loss of coverage is concentrated at the points the optimizer selects (Figure~\ref{fig:fcs-collapse}(a)). Random inputs are covered close to the target in all 30 runs, at $0.895$ to $0.990$. At the returned decisions, coverage falls below $0.90$ in 26 of the 30 runs, and drops as low as $0.17$. Moreover, in almost every run, the returned decisions are covered less than the matched-boundary points, by $0.12$ to $0.70$, even though the matched-boundary points share the returned decisions' distance to the boundary.

We conduct another diagnostic, where we add a penalty to each objective, with weight $\lambda$, that pulls the solver toward a fixed point where $h$ is below zero (Figure~\ref{fig:fcs-collapse}(b)). Here, we see that the coverage returns when the objective no longer drives the solver to the boundary. As $\lambda$ grows from $0$ to $1000$, coverage at the returned decision rises and reaches $1.00$ in 28 of the 30 runs. These results show that the loss of coverage is dependent on how the optimizer converges its search to find the optimal solution for the objective.

Calibrating at a smaller $\alpha$ does not solve this problem (Table~\ref{tab:alpha-sweep}). While a smaller $\alpha$ might hold coverage at the returned decision, it makes their optimality more conservative. For example, at $\alpha=0.01$ the symbolic surrogate finally covers its returned decisions, but their objective values are $19.9\%$ to $25.7\%$ worse than at $\alpha=0.10$. Moreover, the MLP still falls short, by $0.03$ to $0.23$.

\subsection{Certified margins stay within the target risk}
\label{sec:exp-cert}

This study runs SRC on the methane and Williams–Otto benchmarks, with eight seeds each. The candidate margins are those of \eqref{eq:cand-margin} with five shapes, $\theta\in\{0,0.25,0.5,0.75,1\}$. $\theta=0$ is a constant margin and $\theta=1$ a margin proportional to C-MICL's estimate of the surrogate's error. The cost of a margin is the objective value its returned decisions give up compared with the plain conformal deployment $\hat h+\hat q\le 0$, measured as a fraction of each objective's range over the variable box and averaged over objectives. Here, margins are compared at equal cost. The target costs, which we call anchors, lie inside the range of costs that all five shapes can reach: three per seed on methane and two on Williams--Otto, which gives 40 margins in total. At each anchor, SRC sizes every shape to the anchor's cost, selects the shape with the fewest violations on 300 held-out objectives, and certifies the selected margin on 300 new ones. Specifically, with $\gamma=0.12$, $\delta=0.05$ and
$K=40$, a margin is certified when at most 19 of its 300 certification objectives end in a violation. To check the certificates, we solve every selected margin on 200 unseen objectives each.

\begin{figure}[t]
  \centering
  \includegraphics[width=0.55\textwidth]{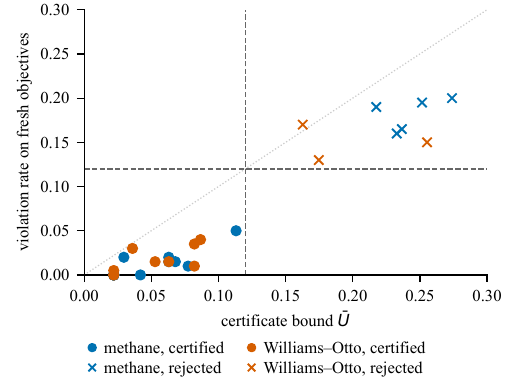}
  \caption{Certification bound $\bar U$ against the violation rate on fresh objectives, for the 30 margins with a fresh evaluation. Dashed lines mark $\gamma=0.12$ and the dotted line is $\bar U$ equal to the fresh rate. No
  certified margin (circle) lies above $\gamma$, but every rejected margin
  (cross) does.}
  \label{fig:cert-bound}
\end{figure}

The results show that the SRC certificate holds (Figure~\ref{fig:cert-bound}). None of the 22 certified margins with an unseen evaluation has a violation rate above $\gamma=0.12$. Precisely, the highest violation rate observed is $0.050$. For all 22, the $95\%$ Clopper--Pearson interval of the rate lies below $\gamma$, with a largest upper end of $0.090$. The 8 rejected margins with a fresh evaluation all lie above $\gamma$, between $0.13$ and $0.20$. Hence, the certificate only rejected the margins that were unsafe. As a stability check, certifying all 40 margins a second time, on 300 new objectives each, gives the same decision for 39 of them.

\subsection{Selected risk at matched cost}
\label{sec:exp-cost}

As we mentioned before, since Theorem~\ref{thm:t4} holds for any locked correction, the shape of the correction is a design choice (Section~\ref{sec:method}). Section~\ref{sec:exp-cert} used the simple family $m_{\theta,\eta}$, which has few candidates and is cheap to choose from. Here we ask how much a richer correction, built directly from the errors at returned decisions, gains over
existing methods when held to the same cost. This
correction adds a term built mainly from $E(x)$ to the conformal constraint, a smooth function fitted to the remaining conformal errors so that it lies above about $90\%$ of them (Appendix~\ref{app:matched} gives the full form). 

In this section, we evaluate SRC’s corrections without the certification step. The certification step only serves to accept or reject a margin after it is locked and does not influence the decisions that margin returns. Hence, certifying it would leave every violation rate in Table 1 unchanged. Since Theorem~\ref{thm:t4} holds for any locked correction, the correction and its certificate can be evaluated separately: Section ~\ref{sec:exp-cert} tests the certificate, and this section tests the correction against baselines that carry no certificate themselves. Wherever a violation rate exceeds the risk target, the full SRC pipeline would abstain, a warning that none of the baselines provides.

We compare it with C-MICL and its ensemble version W-MICL \cite{bib12}, a locally weighted conformal margin \cite{bib31}, scalar padding, which adds a
constant to $\hat q$, $\alpha$-tightening, which recomputes $\hat q$ at a smaller $\alpha$, and a trust region that keeps the decision inside a box
around the training inputs. We also include a C-MICL refit whose uncertainty network is retrained with the same 2,100 returned and boundary points that SRC's correction uses, so that no gain can be put down to extra data alone. The box and the refit are our own controls and not published baselines. The budgets are $0.25$, $0.55$ and $1$ times $C^{\star}$, the average cost of C-MICL at its
published setting, so at the largest budget every method spends what C-MICL spends. Each method's knob is tuned until its cost is within $2\%$ of the
budget. Constraint violations are counted on 200 objectives per seed, over five seeds. When a method cannot be matched on a seed, we still compare it if
SRC's correction cost no more, which favors the baseline (Appendix~\ref{app:matched}). 

\begin{table}[!htbp]
\centering
\footnotesize
\setlength{\tabcolsep}{3pt}
\caption{Violation rates at matched cost (lower is better). Each rate is averaged
over the seeds on which the method could be compared with SRC's correction; rates in
italics use fewer than five seeds. Bold marks the lowest rate in each column. Budgets are fractions of C-MICL's own cost
$C^{\star}$.
$^{\ddagger}$Our controls.}
\label{tab:matched}
\begin{tabular}{l ccc ccc ccc}
\toprule
 & \multicolumn{3}{c}{Methane} & \multicolumn{3}{c}{Williams--Otto} & \multicolumn{3}{c}{Hanoi} \\
\cmidrule(lr){2-4}\cmidrule(lr){5-7}\cmidrule(lr){8-10}
budget $/C^{\star}$ & 0.25 & 0.55 & 1.0 & 0.25 & 0.55 & 1.0 & 0.25 & 0.55 & 1.0 \\
\midrule
SRC correction$^{\ast}$ & 0.42 & \textbf{0.27} & \textbf{0.07} & \textbf{0.18} & \textbf{0.08} & \textbf{0.08} & \textbf{0.35} & \textbf{0.20} & 0.10 \\
\midrule
C-MICL~\cite{bib12} & 0.48 & 0.42 & \textit{0.33} & 0.36 & 0.33 & 0.34 & \textit{0.42} & 0.43 & \textit{0.44} \\
C-MICL refit$^{\ddagger}$ & \textit{0.49} & 0.31 & 0.12 & 0.36 & 0.35 & 0.23 & 0.43 & 0.42 & 0.43 \\
W-MICL~\cite{bib12} & \textit{0.47} & \textit{0.35} & 0.27 & \textit{0.34} & 0.25 & 0.20 & \textit{0.43} & \textit{0.22} & \textit{\textbf{0.07}} \\
Locally weighted~\cite{bib31} & \textit{0.52} & \textit{0.47} & \textit{0.38} & \textit{0.31} & \textit{0.23} & \textit{0.21} & \textbf{0.35} & \textbf{0.20} & 0.08 \\
Scalar padding & 0.44 & 0.32 & 0.26 & 0.34 & 0.26 & 0.20 & 0.41 & 0.32 & 0.22 \\
$\alpha$-tightening & \textit{0.41} & \textbf{0.27} & \textit{0.29} & \textit{0.29} & \textit{0.23} & \textit{0.20} & \textit{0.37} & -- & \textit{0.08} \\
Box trust region$^{\ddagger}$ & \textit{\textbf{0.39}} & 0.31 & 0.28 & 0.26 & 0.26 & 0.26 & \textit{0.38} & 0.31 & 0.20 \\
\bottomrule
\end{tabular}
\end{table}

SRC's correction has the lower violation rate in almost every comparison (Table~\ref{tab:matched}). Among the methods compared on all five seeds, it
has the lowest violation rate in six of the nine columns, and on Hanoi it is level with the locally weighted margin at the two smaller budgets. Compared
with each baseline on the same seeds and objectives, it has fewer or equal violations in 57 of the 62 comparisons and significantly fewer in 53 (exact McNemar test, $p<0.05$). 

On Hanoi, methods like locally weighted margin and $\alpha$-tightening match or beat SRC's correction at some budgets. One possible reason is that Hanoi's surrogate is far more accurate than the other two, with an $R^2$ on its training data of $0.995$ to $0.996$ against $0.76$ to $0.90$. Hence, the errors left after the conformal margin are then small, and a simple local rescaling of the margin captures most of what a fitted correction can add.

The trust region of OptiCL \cite{bib22}, which keeps the decision inside the convex hull of the training inputs, is not in Table~\ref{tab:matched} because we could not compare it at equal costs. Its cheapest setting already costs $1.03$ to $4.77$ times the largest budget on every seed of methane and Williams--Otto. Compared with SRC's correction at the most expensive setting, SRC's correction has fewer violations at 24 of the hull's 30 settings,
significantly at 22. The hull has significantly fewer at 3, all on one methane seed, where SRC's correction could only be compared at less than
half the hull's cost (Appendix~\ref{app:matched}).

\section{Limitations}

 SRC evaluates the true constraint at the decisions the solver returns, and hence, spends oracle calls that could otherwise go into fitting the surrogate. For instance, at one anchor, SRC needed 1,800 oracle calls: 1,500 to choose among the five shapes on 300 objectives, and 300 to certify the locked margin. Moreover, the certificate bounds the selected risk for objectives drawn from $Q$. Our $Q$ is an equal mixture of five families, and hence, might not be a good representation of how objectives arise in any particular application. A certificate for the mixture also says little about each family, and certifying each family separately would require more oracle calls. 

The certificate itself concerns the pipeline as a whole, including the solver and its starting points. A different solver, or the same solver with a different number of starts, can return different decisions which would need to be certified again. Moreover, in situations where every target cost fails, SRC simply returns no margin. While an abstention is the correct outcome when no candidate is safe enough, it leaves the user without a decision rule at that cost. 

Lastly, both the SRC studies use only quantile surrogates fitted with the pinball loss. Such a surrogate is fitted to overestimate the true constraint at most of its training points. Hence, a violation of the true constraint at the returned decision cannot be blamed on a surrogate that is optimistic everywhere. While the guarantee of Theorem~\ref{thm:t4} does not depend on the type of surrogate, how often SRC certifies a correction, and at what cost, may. We have not explored how SRC behaves on other surrogate types.

\section{Conclusion}

This study examines how a surrogate constraint can be calibrated to ensure the feasibility of an optimizer's returned decision. We showed that coverage at random points does not by itself bound the feasibility of that decision, and the likelihood-ratio reweighting used for feedback covariate shift does not apply when the constraint is active there. Our method, selected-risk calibration (SRC), calibrates the constraint on the decisions the solver returns and certifies the result with a finite-sample bound that holds for all certified margins at once. On our benchmarks, a margin covering about $90\%$ of random inputs covered as few as $17\%$ of the returned decisions, while none of the certified margins we checked exceeded its risk target. Moreover, at equal cost, SRC's correction had a lower or equal violation rate than the baselines in 57 of 62 comparisons. The certificate uses nothing specific to constraint learning and applies wherever a learned model is used to choose a decision that can later be checked. Future work could explore other surrogate families, cheaper ways to choose the shape of the correction, and certificates that hold for each family of objectives separately.

\bibliographystyle{unsrtnat}
\bibliography{references}  






\appendix

\clearpage

\section{Proof of Theorem~\ref{thm:separation} and further remarks}\label{app:t1}
 
\begin{proof}[Proof of Theorem~\ref{thm:separation}]
Let $\mathcal{X}=[0,1]^{2}$ with points $x=(u,v)$, let $P$ be the uniform
distribution on $\mathcal{X}$, and let
\[
  \hat{h}(u,v)=v-1,
  \qquad
  h(u,v)=u+v-1 .
\]
The surrogate does not involve the coordinate $u$, on which the true constraint $h$ depends. Here, the
residual $s(u,v)$ is $u$. The objectives are $c_{\beta}(u,v)=\beta u-v$ with $\beta\in[-1,1]$, and
$Q$ is any distribution of $\beta$ with $Q(\beta<0)=p$ and $Q(\beta=0)=0$. For instance, $\beta=-1$ with probability $p$ and $\beta=1$ otherwise. The term $-v$ rewards pushing against the learned constraint, and the term $\beta u$ expresses a preference along it. Note that none of these choices involves $\alpha$.
 
Under $P$ the coordinate $u$ is uniform on $[0,1]$, so $S_{1},\dots,S_{m+1}$ are i.i.d.\ uniform on $[0,1]$ and almost surely distinct. The rank of $S_{m+1}$ among $S_{1},\dots,S_{m+1}$ is then uniform on $\{1,\dots,m+1\}$, and $S_{m+1}\le S_{(k)}$ when this rank is at most $k$. Hence, the coverage is $k/(m+1)$. Part~\ref{it:cov} follows from
$(1-\alpha)(m+1)\le k<(1-\alpha)(m+1)+1$.
We also record that $\hat{q}=S_{(k)}$ lies in $[0,1]$, and that $\hat{q}<1$ with probability one, since $\hat{q}\le\max_{i}S_{i}$ and $\Pr(\max_{i}S_{i}=1)=0$.
 
With $g\equiv 0$,
\[
  \widehat{\mathcal{X}}(0)
  =\bigl\{(u,v)\in[0,1]^{2}: v-1+\hat{q}\le 0\bigr\}
  =[0,1]\times[0,1-\hat{q}],
\]
which is non-empty because $\hat{q}\le 1$.
The set is a rectangle. Moreover, $c_{\beta}$ is the sum of a function of $u$ and a function of $v$, and hence, the two coordinates can be minimized
separately. The term $-v$ is minimized only at $v^{\star}=1-\hat{q}$, irrespective of the value of $\beta$. The term $\beta u$ is minimized at
$u^{\star}=1$ if $\beta<0$, and at $u^{\star}=0$ if $\beta>0$. The minimizer is therefore unique whenever $\beta\neq 0$, which holds for
$Q$-almost every objective.
 
Substituting
$v^{\star}=1-\hat{q}$ gives
\[
  h(x^{\star})=u^{\star}+v^{\star}-1=u^{\star}-\hat{q}.
\]
If $\beta<0$ this equals $1-\hat{q}>0$, and if $\beta>0$ it equals $-\hat{q}\le 0$. Hence, the returned decision is infeasible exactly when $\beta<0$. Therefore, $R(0)=Q(\beta<0)=p$ when $\hat{q}<1$. This proves part~\ref{it:risk}.
\end{proof}
 
The identity $h(x^{\star})=u^{\star}-\hat{q}$ shows where the argument turns. The margin $\hat{q}$ is determined by the distribution of the
residual at points drawn from $P$. However, the residual at the returned decision is $u^{\star}$, whose value is fixed by the objective. The
strip $\{u>\hat{q}\}$, on which the padded surrogate still under-predicts $h$, has probability $1-\hat{q}$ under $P$, which is close to $\alpha$ for
large $m$. A larger calibration sample only moves $\hat{q}$ closer to $1-\alpha$, but does not change where the solver goes.
 
\begin{remark}[No bound in either direction]\label{rem:directions} Theorem~\ref{thm:separation} rules out any bound on the selected risk in terms of the level. Suppose that $R(0)\le f(\alpha)$ held for every
instance, for some function $f$ with $f(\alpha)\to 0$ as $\alpha\to 0$. At every level $\alpha$ and for
$m$ large enough that $\alpha>1/(m+1)$, taking any $p>0$ in the theorem gives an instance with $R(0)=p$. This contradicts the bound as soon as $\alpha$ is small enough that $f(\alpha)<p$. The theorem equally allows $p<\alpha$. Hence, the selected risk can also fall below the nominal level. \end{remark}
 
\begin{remark}[The surrogate need not be biased]\label{rem:unbiased}
In the construction the surrogate satisfies $\hat{h}\le h$ everywhere, hence,
it never errs on the safe side. However, this does not drive the result. Since $u$ is independent of $v$ under $P$ and has mean $\tfrac12$, the least-squares fit of $h$ on $v$ alone is
$\hat{h}(u,v)=v-\tfrac12$. With this surrogate the
residual $s=u-\tfrac12$ is uniform on $[-\tfrac12,\tfrac12]$, the
deployed set is $[0,1]\times[0,\tfrac12-\hat{q}]$, and the argument above gives $h(x^{\star})=u^{\star}-\tfrac12-\hat{q}$. With probability one this
is positive when $\beta<0$ and negative when $\beta>0$, so, with the residual now uniform on $[-\tfrac12,\tfrac12]$ in place of $[0,1]$, the coverage is still $k/(m+1)$ by the same rank argument and $R(0)=p$ as before.
\end{remark}

 \FloatBarrier
\section{Proof of Proposition~\ref{thm:t2}}\label{app:t2}
 
\begin{proof}[Proof of Proposition~\ref{thm:t2}]
We may treat $\hat{h}$ and $g$ as fixed since they do not depend on $\mathcal{D}_{\mathrm{cal}}$. Let $\phi=\hat{h}+g$ and, for
$t\in\mathbb{R}$, let $M_{t}=\{x\in\mathcal{X}:\phi(x)=-t\}$. Now, $M=M_{\hat{q}}$.
 
\emph{$P$ gives $M$ probability zero.} The sets $M_{t}$ are pairwise disjoint, because each point $x$ has a single value $\phi(x)$. For each
$n\ge 1$, fewer than $n$ of them can have $P(M_{t})>1/n$. Otherwise, their probabilities would add up to more than one. The set
$T=\{t:P(M_{t})>0\}$ is therefore countable and determined by $P$ and $\phi$. However, the margin $\hat{q}$, is always one of $S_{1},\dots,S_{m}$. Hence, for every $t$
\[
  \Pr(\hat{q}=t)\le\sum_{i=1}^{m}\Pr(S_{i}=t)=0,
\]
since the residuals have a continuous distribution. Summing over the countably many elements of $T$ gives $\Pr(\hat{q}\in T)=0$. Hence, with
probability one, $P(M)=P(M_{\hat{q}})=0$.
 
\emph{$\tilde P$ gives $M$ probability one.} Fix a calibration sample with $P(M)=0$ and a non-empty deployed set. $\phi$ is continuous and $x^{\star}$ is measurable. Hence, the set $M$ is closed. Moreover, $\{c:x^{\star}(c)\in M\}$ is an event. By the active-constraint assumption, it has $Q$-probability one, that is, $\tilde P(M)=1$.
 
\emph{Conclusion.} The set $M$ has $P(M)=0$ and $\tilde P(M)=1$. Hence, $P$ and $\tilde P$ are mutually singular, and $\tilde P$ is not absolutely
continuous with respect to $P$. In particular no likelihood ratio exists: a function $f$ with $\tilde P(A)=\int_{A}f\,dP$ for every $A$ would give
$\tilde P(M)=\int_{M}f\,dP=0$. The coverage theorem of \cite{bib13} assumes absolute continuity for every possible data set, and hence, does not
apply. \end{proof}
 
\begin{remark}[When the conclusion holds for every calibration sample]
\label{rem:every-sample}
The calibration sample enters the proof only to rule out the countably many exceptional values of $\hat{q}$. If $P$ has a density and $M$ has
zero volume, then $P(M)=0$, and the conclusion holds for that calibration sample. The level set $M$ has zero volume in two common cases. If $\phi=\hat{h}+g$ is continuously differentiable near $M$ with a
non-vanishing gradient on $M$, the implicit function theorem shows that $M$ is locally the graph of a function of $d-1$ coordinates, and such
graphs have zero volume. If $\phi$ is continuous and piecewise affine with finitely many pieces, then on each piece $M$ is empty, contained in a hyperplane, or equal to the whole piece, and the last case requires $\phi$ to equal $-\hat{q}$ on a piece with non-empty interior.
\end{remark}

 \FloatBarrier
\section{Proof of Theorem~\ref{thm:t4}}\label{app:t4}
 
Fix $n$ and $\varepsilon\in(0,1)$. For $r\in[0,1]$ and $v\in\{0,\dots,n\}$ let $F_{r}(v)=\Pr\bigl(\mathrm{Bin}(n,r)\le v\bigr)$. Hence,
$U_{n}(v;\varepsilon)=\sup\{r\in[0,1]:F_{r}(v)\ge\varepsilon\}$. The set in this supremum is never empty, because $F_{0}(v)=1$. Two facts follow from the definition:
\begin{enumerate}
\item[(F1)] If $r>U_{n}(v;\varepsilon)$, then $F_{r}(v)<\varepsilon$, since otherwise $r$ would belong to the set whose supremum is $U_{n}(v;\varepsilon)$.
\item[(F2)] $U_{n}(v;\varepsilon)$ is non-decreasing in $v$, since $F_{r}(v)$ is non-decreasing in $v$ for every $r$, so the set grows with $v$.
\end{enumerate}

For computation, note that for $v<n$ the map $r\mapsto F_{r}(v)$ is continuous and strictly decreasing from $1$ to $0$, so $U_{n}(v;\varepsilon)$ is the unique $r$ with $F_{r}(v)=\varepsilon$. The identity $F_{r}(v)=1-I_{r}(v+1,n-v)$, where $I$ is the regularized
incomplete beta function, identifies this $r$ with the $(1-\varepsilon)$-quantile of $\mathrm{Beta}(v+1,n-v)$. The proof does not use this. However, the proof does use a comparison due to Hoeffding.
 
\begin{lemma}[Hoeffding~\cite{bib33}]\label{lem:hoeffding} Let $N$ be the sum of $n$ independent $\{0,1\}$-valued random variables with success probabilities $p_{1},\dots,p_{n}$, and let $\bar{p}=\frac{1}{n}\sum_{j}p_{j}$. If $b$ is an integer with
$0\le b\le n\bar{p}-1$, then $\Pr(N\le b)\le\Pr\bigl(\mathrm{Bin}(n,\bar{p})\le b\bigr)$.
\end{lemma}
 
Put simply, among all sums of $n$ independent trials with the same average success probability, the binomial one, whose trials all have the same
probability, is the most likely to fall to $b$ or below, for every $b$ at least one below the mean.
 
\begin{proof}[Proof of Theorem~\ref{thm:t4}]
Fix a margin and a tolerance $\tau\ge 0$, drop the index $k$, and condition on the data used to build $g$. We write $R(g)$  for $R_\tau(g)$ throughout this proof. Given this data, $g$ and $Q_{1},\dots,Q_{L}$ are fixed, so $R(g)$ is a fixed number, and the certification objectives $c_{1},\dots,c_{n}$ are independent, with $n/L$ of them drawn from each $Q_{\ell}$.
 
\emph{Step 1} Let $V_{j}=1$ if the solve of $c_{j}$ under the deployed constraint $\hat{h}+\hat{q}+g\le 0$ fails or returns a point $x$ with $h(x)> \tau$. Otherwise $V_{j}=0$ so that $N=\sum_{j}V_{j}$. The
solver is deterministic given its starting points, so $V_{j}$ is a function of $c_{j}$, and $V_{1},\dots,V_{n}$ are independent. If $c_{j}$ is drawn from $Q_{\ell}$, then $\Pr(V_{j}=1)$ is
the selected risk of $g$ under $Q_{\ell}$, at tolerance $\tau$,  which we write $R_{\ell}(g)$. Each family contributes $n/L$ objectives, so the average success probability is
\[
  \bar{p}=\frac{1}{n}\sum_{j=1}^{n}\Pr(V_{j}=1)
  =\frac{1}{L}\sum_{\ell=1}^{L}R_{\ell}(g)=R(g),
\]
where the last equality holds because $Q$ is the equal mixture of the $Q_{\ell}$.
 
\emph{Step 2} The margin is certified when $U_{n}(N;\varepsilon)\le\gamma$. By (F2) this happens exactly when $N\le b$. Moreover, it never happens if $b=-1$.
 
\emph{Step 3} If $R(g)\le\gamma$ or $b=-1$, the event in the theorem cannot occur. Otherwise $R(g)>\gamma$ and $b\ge 0$, and the assumption on $b$ gives
\[
  b\le n\gamma-1<nR(g)-1 .
\]
Lemma~\ref{lem:hoeffding} with $\bar{p}=R(g)$ therefore applies, and
\[
  \Pr(N\le b)\le F_{R(g)}(b)<\varepsilon ,
\]
where the last inequality is (F1), because
$R(g)>\gamma\ge U_{n}(b;\varepsilon)$. In every case, then, the probability given the data that the margin is certified with $R(g)>\gamma$ is at most $\varepsilon$, and averaging over the data gives the first claim.
 
\emph{Step 4} Let $E_{k}$ be the event that margin $k$ is certified and $R(g_{k})>\gamma$. For any events, $\Pr(E_{1}\cup\dots\cup  E_{K})\le\sum_{k}\Pr(E_{k})$. These events may depend on one another, through shared certification objectives, but the union bound needs no independence, so the probability that some margin is certified with selected risk above $\gamma$ is at most $K\varepsilon\le\delta$.
\end{proof}
 
\begin{remark}[Independent draws from $Q$]\label{rem:iid} The comparison of Lemma~\ref{lem:hoeffding} is needed because our
certification sets are stratified by family. When the certification objectives are independent draws from $Q$ itself, as when $L=1$, the count $N$ is binomial with parameter $\rho=R(g)$ given
the data, and Clopper--Pearson's classical guarantee
$\Pr\bigl(\rho>U_{n}(N;\varepsilon)\bigr)\le\varepsilon$ holds with no
condition on $b$. By (F1) the event $\{\rho>U_{n}(N;\varepsilon)\}$ is contained in $\{F_{\rho}(N)<\varepsilon\}$. Since $F_{\rho}$ is
non-decreasing, this larger event is $\{N\le v^{\ast}\}$, where $v^{\ast}$ is the largest $v$ with $F_{\rho}(v)<\varepsilon$ (the event is empty if
there is none). Its probability is $F_{\rho}(v^{\ast})<\varepsilon$.
\end{remark}
 
\begin{remark}[Future objectives from a different distribution]
\label{rem:shift}
If future objectives follow $Q'$ instead of $Q$, then, writing $R_{Q}$ and
$R_{Q'}$ for the selected risk under each,
$R_{Q'}(g)\le R_{Q}(g)+d_{\mathrm{TV}}(Q,Q')$ for every correction $g$, because both risks are the probability of the same set of objectives.
Consequently, when $K\varepsilon\le\delta$, outside an event of probability at most $\delta$, every certified margin has selected risk at most
$\gamma+d_{\mathrm{TV}}(Q,Q')$ under $Q'$.
\end{remark}

 \FloatBarrier
\section{The coverage diagnostic}
\label{app:fcs}

\paragraph{No margin against the conformal margin.}
Table~\ref{tab:nomargin} gives, for every benchmark and seed, the number of violations with no margin and with the conformal margin, and the margin's
coverage of random held-out inputs. These runs use the surrogates of the matched-cost comparison (Section~\ref{sec:exp-cost}). In 6 of the 15 runs
the conformal margin gives slightly more violations than no margin at all. 

\begin{table}[!htbp]
\centering
\small
\caption{The surrogate deployed without a margin ($\hat h\le 0$) and with the
conformal margin ($\hat h+\hat q\le 0$): violations among 200 objectives per seed,
and the margin's coverage of 800 random held-out inputs. $^{\dagger}$Runs where the
margin gives more violations than no margin.}
\label{tab:nomargin}
\begin{tabular}{llcccc}
\toprule
benchmark & seed & $\hat q$ & no margin & conformal margin & coverage \\
\midrule
Methane & 3 & $+0.374$ & 106/200 & 97/200 & 0.940 \\
 & 5 & $-0.083$ & 71/200 & 74/200$^{\dagger}$ & 0.889 \\
 & 6 & $+0.071$ & 104/200 & 105/200$^{\dagger}$ & 0.914 \\
 & 7 & $+0.175$ & 109/200 & 110/200$^{\dagger}$ & 0.921 \\
 & 9 & $+0.340$ & 98/200 & 99/200$^{\dagger}$ & 0.906 \\
\addlinespace[2pt]
Williams--Otto & 3 & $+0.008$ & 72/200 & 69/200 & 0.914 \\
 & 5 & $-0.001$ & 70/200 & 71/200$^{\dagger}$ & 0.904 \\
 & 6 & $+0.003$ & 66/200 & 64/200 & 0.923 \\
 & 7 & $+0.006$ & 67/200 & 69/200$^{\dagger}$ & 0.905 \\
 & 9 & $+0.007$ & 91/200 & 79/200 & 0.918 \\
\addlinespace[2pt]
Hanoi & 3 & $+1.991$ & 73/200 & 69/200 & 0.918 \\
 & 5 & $+1.724$ & 114/200 & 107/200 & 0.906 \\
 & 6 & $+1.869$ & 103/200 & 102/200 & 0.896 \\
 & 7 & $+2.738$ & 128/200 & 120/200 & 0.913 \\
 & 9 & $+0.943$ & 89/200 & 87/200 & 0.898 \\
\bottomrule
\end{tabular}
\end{table}

\paragraph{Set-up of the diagnostic.}
The diagnostic is a separate experiment, with its own training, calibration and test splits and its own margins, so its numbers should not
be compared with those of the other studies. Each repetition draws 100 linear objectives in the way C-MICL does \cite{bib12}: every weight is
uniform on $[-4,4]$, and negative weights are divided by 10. Each objective is solved with SLSQP from 12 starting points. The symbolic surrogate is an
expression found with LLM-SR \cite{bib35} and is the same in every repetition. The MLP surrogate is a multilayer perceptron with two hidden layers of 32 units, trained with the squared loss and retrained
in every repetition. The scalar margin is the conformal quantile of the signed residuals $h-\hat h$ on the calibration split. For the normalized
margin, a second network of the same size is trained to predict $|h(x)-\hat h(x)|$. Its prediction $\hat u(x)$ is floored at one fifth of
the median absolute training residual, and the margin is $m(x)=\hat q_{\mathrm{n}}\,\hat u(x)$, where $\hat q_{\mathrm{n}}$ is the
conformal quantile of $|h-\hat h|/\hat u$ on the calibration split. This network is also retrained in every repetition. The random inputs are the
200 points of the test split, which does not overlap the training and calibration splits.

\paragraph{Matched-boundary points.}
We draw points uniformly from the variable box and keep those that satisfy the benchmark's operating constraints and whose value of
$\hat h(x)+m(x)$ lies between the 5th and 95th percentiles of its values at the returned decisions; the lower end of this band is set to at most
$-10^{-3}$, so that points just inside the deployed set are included. Sampling stops when 400 points have been kept or 200,000 drawn. For methane
with the symbolic surrogate and the normalized margin, the band is about $10^{-3}$ wide, against 5 to 6 in the other configurations, and in
three of the five repetitions no point was kept, so the comparison with the returned decisions cannot be made there. In this configuration the returned
decisions are covered at $0.94$ to $1.00$ in four repetitions and at $0.80$ in the fifth. Table~\ref{tab:fcs-fits} summarizes each configuration; the results of every run are in the code repository.

\begin{table}[!htbp]
\centering
\footnotesize
\setlength{\tabcolsep}{4pt}
\caption{The coverage diagnostic, summarized over the five repetitions of each
configuration. Where matched-boundary points were found in fewer than five repetitions, the number is given in brackets.}
\label{tab:fcs-fits}
\begin{tabular}{lccccc}
\toprule
 & random & returned & below & matched & \\
configuration & inputs & decisions & $0.90$ & boundary & drop \\
\midrule
Methane / symbolic / scalar & 0.910 & 0.53--0.71 & 5 of 5 & 0.875--0.920 & 0.205--0.353 \\
Methane / MLP / scalar & 0.955--0.990 & 0.19--0.60 & 5 of 5 & 0.855--0.990 & 0.325--0.703 \\
Methane / MLP / normalized & 0.955--0.990 & 0.17--0.46 & 5 of 5 & 0.790--0.933 & 0.330--0.703 \\
Methane / symbolic / normalized & 0.980--0.990 & 0.80--1.00 & 1 of 5 & 0.987--1.000 (2 of 5) & 0.000--0.187 \\
Williams--Otto / symbolic / scalar & 0.895 & 0.74--0.78 & 5 of 5 & 0.993--1.000 & 0.220--0.255 \\
Williams--Otto / MLP / scalar & 0.895--0.940 & 0.53--0.88 & 5 of 5 & 0.993--1.000 & 0.120--0.463 \\
\bottomrule
\end{tabular}
\end{table}

\paragraph{Dose-response.}
The penalized objective is
\[
  c_{\lambda}(x)=c(x)+\frac{\lambda}{d}\sum_{j=1}^{d}
  \Bigl(\frac{x_{j}-x_{\mathrm{safe},j}}{r_{j}}\Bigr)^{2},
\]
where $d$ is the number of variables, $r_{j}$ is the width of the range of variable $j$, and $x_{\mathrm{safe}}$ is the average of the training inputs whose value of $h$ is among the lowest $15\%$. It depends only on the training split, so it is the same point in every repetition. It has $h(x_{\mathrm{safe}})=-4.69$ on methane and $-1.22$ on Williams--Otto and lies in the variable box. On methane it misses the operating constraint $v_0\le 1.1\,T$ by $99.6$\,cm$^3$/h, since the training inputs are drawn uniformly from the box. We use $\lambda=0$, 1, 3, 10, 30, 100, 300 and 1000. 

\paragraph{Smaller $\alpha$.}
The $\alpha$ sweep repeats the diagnostic on methane with the scalar margin at $\alpha=0.20$, $0.15$, $0.10$, $0.05$, $0.02$ and $0.01$, for
both surrogates and the same five repetitions. Coverage of random inputs is never more than $0.06$ below its target. At $\alpha=0.01$ the objective
values of the returned decisions are $19.9\%$ to $25.7\%$ worse than at $\alpha=0.10$ with the symbolic surrogate, and $3.3\%$ to $11.8\%$ worse
with the MLP. Table~\ref{tab:alpha-sweep} gives the minimum and maximum over repetitions of every quantity.

\begin{table}[!htbp]
\centering
\small
\caption{The $\alpha$ sweep on methane over five repetitions, reported as the minimum and maximum across them. The gap is $(1-\alpha)$ minus coverage at $x^{\star}$, so a positive gap is coverage promised but not delivered, and the mean objective is minimised, so a larger value is worse.}
\label{tab:alpha-sweep}
\begin{tabular}{l r rr rr rr rr}
\toprule
surrogate & $\alpha$ & \multicolumn{2}{c}{random inputs} & \multicolumn{2}{c}{at $x^{\star}$} & \multicolumn{2}{c}{gap} & \multicolumn{2}{c}{mean objective} \\
\cmidrule(lr){3-4}\cmidrule(lr){5-6}\cmidrule(lr){7-8}\cmidrule(lr){9-10}
 & & min & max & min & max & min & max & min & max \\
\midrule
symbolic & 0.20 & 0.740 & 0.740 & 0.500 & 0.680 & +0.120 & +0.300 & 22.789 & 32.398 \\
 & 0.15 & 0.835 & 0.835 & 0.530 & 0.700 & +0.150 & +0.320 & 23.043 & 32.656 \\
 & 0.10 & 0.910 & 0.910 & 0.530 & 0.710 & +0.190 & +0.370 & 23.355 & 32.971 \\
 & 0.05 & 0.955 & 0.955 & 0.580 & 0.780 & +0.170 & +0.370 & 23.716 & 33.335 \\
 & 0.02 & 0.965 & 0.965 & 0.630 & 0.800 & +0.180 & +0.350 & 24.292 & 33.915 \\
 & 0.01 & 0.990 & 0.990 & 1.000 & 1.000 & -0.010 & -0.010 & 29.364 & 39.544 \\
\midrule
MLP & 0.20 & 0.905 & 0.960 & 0.150 & 0.510 & +0.290 & +0.650 & 21.500 & 30.336 \\
 & 0.15 & 0.935 & 0.980 & 0.140 & 0.550 & +0.300 & +0.710 & 21.734 & 30.512 \\
 & 0.10 & 0.955 & 0.990 & 0.190 & 0.600 & +0.300 & +0.710 & 21.934 & 30.686 \\
 & 0.05 & 0.960 & 1.000 & 0.460 & 0.740 & +0.210 & +0.490 & 22.301 & 31.053 \\
 & 0.02 & 0.985 & 1.000 & 0.460 & 0.840 & +0.140 & +0.520 & 23.394 & 31.383 \\
 & 0.01 & 0.985 & 1.000 & 0.760 & 0.960 & +0.030 & +0.230 & 23.604 & 31.706 \\
\bottomrule
\end{tabular}
\end{table}

\FloatBarrier
\section{Experimental details}
\label{app:details}

Table~\ref{tab:benchmarks} lists the decision variables and constraints of
each benchmark. The methane reactor, its variables and its operating constraints are those of the C-MICL regression case study \cite{bib12}; the
simulator integrates the reactor model with SciPy's \texttt{solve\_ivp} at
tolerances $10^{-10}$. The Williams--Otto simulator solves the recycle loop
with SciPy's \texttt{fsolve}. The Hanoi
network has 31 demand junctions, a reservoir at a fixed head of $100$\,m and 34 pipes, and is simulated in steady state with EPANET~2.2 through the
WNTR package. 

\begin{table}[!htbp]
\centering
\small
\caption{Benchmarks. Each learned constraint is written as $h(x)\le 0$.}
\label{tab:benchmarks}
\begin{tabular}{p{0.16\textwidth}p{0.40\textwidth}p{0.36\textwidth}}
\toprule
benchmark & decision variables and bounds & constraints \\
\midrule
Methane reactor
& methane feed $v_0$ and helium sweep $v_{\mathrm{He}}$, both in
  $[450,1500]$\,cm$^3$/h; temperature $T\in[997.18,1348.12]$\,K; tube
  diameter $d\in[0.5,2]$\,cm; length $L\in[10,100]$\,cm
& learned: benzene outlet flow $\ge 50$; operating: $10d\le L\le 150d$,
  $0.75\,v_{\mathrm{He}}\le v_0\le 3\,v_{\mathrm{He}}$,
  $20L\le v_0\le 120L$, $v_0\le 1.1\,T$ \\
\addlinespace[3pt]
Williams--Otto
& reactor volume $V\in[0.03,0.10]$; scaled temperature $T\in[5.8,6.8]$;
  fresh feeds $F_A,F_B\in[1,5]$; purge fraction $\eta\in[0.1,0.9]$
& learned: product flow $F_P\ge 0.64$ \\
\addlinespace[3pt]
Hanoi network
& eight group diameters $D_g\in[12,40]$\,in; the variables are
  $x_g=D_g^{1.5}\in[41.57,252.98]$, since the published pipe cost is
  linear in $D^{1.5}$
& learned: pressure head at every junction $\ge 30$\,m \\
\bottomrule
\end{tabular}
\end{table}

\paragraph{Objective families.}
Outside the diagnostic, an objective is $c(x)=w^{\top}x$ with $w=W/\|W\|$,
and $W$ is drawn from one of five families. In the first, $W\sim N(0,I)$,
so the direction of $w$ is uniformly random. In the second,
$W=\zeta+0.1\,\xi$, where $\zeta$ is uniform on $\{-1,1\}^{d}$ and
$\xi\sim N(0,I)$, so $w$ points to a corner of the box. In the third,
$W_{j}=\operatorname{sign}(r_{j})\,|r_{j}|^{3}$ with $r\sim N(0,I)$, so a
few variables dominate. The last two families pick a calibration input
$x_{i}$ and set $w$ so that minimizing $c$ drives the solver toward
$x_{i}$, with a random perturbation of scale $0.20$ and $0.12$
respectively. The fourth picks $x_{i}$ among the calibration inputs whose
residual $h-\hat h$ is in the top $10\%$. The fifth picks it among the inputs whose residual and whose
closeness to the boundary of the deployed set, measured by
$-|\hat h(x)+\hat q|$, are both at or above their 60th percentile.

In both the certification study and the matched-cost comparison, the surrogates for methane and
Williams--Otto are expressions found with LLM-SR \cite{bib35} whose coefficients are refitted with the pinball loss, so that $\hat h(x)$ estimates an upper quantile of $h$ at $x$. The two studies start from different expressions on methane. For Hanoi, the matched-cost comparison uses a parametric form motivated by the network's hydraulics. The conformal
margin $\hat q$ is always the one-sided quantile of the calibration residuals at $\alpha=0.10$. 

\FloatBarrier

\section{Certification study}
\label{app:cert}

Each shape is sized to an anchor's cost on 100 matching objectives, to within $1\%$ of the target. The shape is chosen on 300 further objectives,
and each certificate uses 300 objectives, 60 from each family. A returned decision counts as a violation when $h(x^{\star})>10^{-8}$. A solve that
returns no feasible point also counts as a violation. Each objective is solved from 500 starting points. The constants $\gamma=0.12$, $\delta=0.05$, $K=40$ and $n=300$ were decided before any certificate was computed.

At one anchor, SRC needs 1,800 oracle calls: 1,500 to choose among the five shapes on 300 objectives, and 300 to certify the locked margin.

\begin{table}[!htbp]
\centering
\scriptsize
\caption{All 40 margins of the certification study. Each margin is certified twice,
on two independent sets of 300 objectives; ``cert.'' says whether $\bar U\le\gamma=0.12$
at level $\delta/K$ with $K=40$, which holds exactly when $v\le 19$. The last column
counts violations on fresh objectives.}
\label{tab:cert-all}
\begin{tabular}{llcc ccc ccc c}
\toprule
 & & & & \multicolumn{3}{c}{first certification} & \multicolumn{3}{c}{second certification} & \\
\cmidrule(lr){5-7}\cmidrule(lr){8-10}
benchmark & seed & anchor & $\theta$ & $v/300$ & $\bar U$ & cert. & $v/300$ & $\bar U$ & cert. & fresh \\
\midrule
Methane & 3 & 0.25 & 0.00 & 53/300 & 0.252 & no & 54/300 & 0.255 & no & 39/200 \\
 & 3 & 0.50 & 0.00 & 7/300 & 0.063 & yes & 4/300 & 0.047 & yes & 4/200 \\
 & 3 & 0.75 & 0.25 & 7/300 & 0.063 & yes & 4/300 & 0.047 & yes & 3/200 \\
 & 4 & 0.25 & 0.00 & 15/300 & 0.100 & yes & 15/300 & 0.100 & yes & -- \\
 & 4 & 0.50 & 0.00 & 9/300 & 0.073 & yes & 13/300 & 0.091 & yes & -- \\
 & 4 & 0.75 & 0.00 & 9/300 & 0.073 & yes & 12/300 & 0.087 & yes & -- \\
 & 5 & 0.25 & 0.00 & 18/300 & 0.113 & yes & 28/300 & 0.155 & no & 10/200 \\
 & 5 & 0.50 & 0.00 & 8/300 & 0.068 & yes & 6/300 & 0.058 & yes & 3/200 \\
 & 5 & 0.75 & 0.50 & 3/300 & 0.042 & yes & 1/300 & 0.030 & yes & 0/200 \\
 & 6 & 0.25 & 0.00 & 44/300 & 0.218 & no & 42/300 & 0.210 & no & 38/200 \\
 & 6 & 0.50 & 0.00 & 1/300 & 0.030 & yes & 1/300 & 0.030 & yes & 4/200 \\
 & 6 & 0.75 & 0.00 & 0/300 & 0.022 & yes & 0/300 & 0.022 & yes & 0/200 \\
 & 7 & 0.25 & 0.00 & 10/300 & 0.077 & yes & 18/300 & 0.113 & yes & 2/200 \\
 & 7 & 0.50 & 0.00 & 0/300 & 0.022 & yes & 0/300 & 0.022 & yes & 0/200 \\
 & 7 & 0.75 & 0.00 & 0/300 & 0.022 & yes & 0/300 & 0.022 & yes & 0/200 \\
 & 8 & 0.25 & 0.00 & 0/300 & 0.022 & yes & 0/300 & 0.022 & yes & 0/200 \\
 & 8 & 0.50 & 0.00 & 0/300 & 0.022 & yes & 0/300 & 0.022 & yes & 0/200 \\
 & 8 & 0.75 & 0.00 & 0/300 & 0.022 & yes & 0/300 & 0.022 & yes & 0/200 \\
 & 9 & 0.25 & 0.00 & 59/300 & 0.274 & no & 59/300 & 0.274 & no & 40/200 \\
 & 9 & 0.50 & 0.00 & 49/300 & 0.237 & no & 51/300 & 0.244 & no & 33/200 \\
 & 9 & 0.75 & 0.00 & 48/300 & 0.233 & no & 48/300 & 0.233 & no & 32/200 \\
 & 10 & 0.25 & 0.00 & 55/300 & 0.259 & no & 47/300 & 0.229 & no & -- \\
 & 10 & 0.50 & 0.00 & 51/300 & 0.244 & no & 46/300 & 0.225 & no & -- \\
 & 10 & 0.75 & 0.00 & 0/300 & 0.022 & yes & 0/300 & 0.022 & yes & -- \\
\addlinespace[3pt]
Williams--Otto & 3 & 1/3 & 0.75 & 11/300 & 0.082 & yes & 13/300 & 0.091 & yes & 2/200 \\
 & 3 & 2/3 & 0.50 & 0/300 & 0.022 & yes & 0/300 & 0.022 & yes & 0/200 \\
 & 4 & 1/3 & 0.25 & 59/300 & 0.274 & no & 62/300 & 0.285 & no & -- \\
 & 4 & 2/3 & 0.25 & 58/300 & 0.270 & no & 62/300 & 0.285 & no & -- \\
 & 5 & 1/3 & 0.75 & 54/300 & 0.255 & no & 61/300 & 0.281 & no & 30/200 \\
 & 5 & 2/3 & 0.75 & 12/300 & 0.087 & yes & 10/300 & 0.077 & yes & 8/200 \\
 & 6 & 1/3 & 0.50 & 0/300 & 0.022 & yes & 1/300 & 0.030 & yes & 1/200 \\
 & 6 & 2/3 & 0.50 & 0/300 & 0.022 & yes & 0/300 & 0.022 & yes & 0/200 \\
 & 7 & 1/3 & 0.75 & 11/300 & 0.082 & yes & 4/300 & 0.047 & yes & 7/200 \\
 & 7 & 2/3 & 0.50 & 5/300 & 0.053 & yes & 3/300 & 0.042 & yes & 3/200 \\
 & 8 & 1/3 & 0.75 & 33/300 & 0.175 & no & 37/300 & 0.190 & no & 26/200 \\
 & 8 & 2/3 & 1.00 & 7/300 & 0.063 & yes & 15/300 & 0.100 & yes & 3/200 \\
 & 9 & 1/3 & 0.75 & 30/300 & 0.163 & no & 40/300 & 0.202 & no & 34/200 \\
 & 9 & 2/3 & 0.75 & 2/300 & 0.036 & yes & 7/300 & 0.063 & yes & 6/200 \\
 & 10 & 1/3 & 0.75 & 2/300 & 0.036 & yes & 5/300 & 0.053 & yes & -- \\
 & 10 & 2/3 & 0.50 & 1/300 & 0.030 & yes & 3/300 & 0.042 & yes & -- \\
\bottomrule
\end{tabular}
\end{table}

\FloatBarrier
\section{Matched-cost comparison}
\label{app:matched}

\paragraph{SRC's correction.}
The deployed constraint is
\[
  \hat h(x)+\hat q+s\bigl[(1-w)\,E(x)+w\,F\bigr]\le 0,
\]
with $w=0.5$ and the knob $s\ge 0$. $E$ is a smooth maximum of affine functions, fitted with the pinball loss at level $0.90$ to the residuals $h-\hat h-\hat q$, so that it lies above about $90\%$ of the weighted residuals. $F$ is the $0.90$-quantile of the same residuals on the 400 held-out inputs,
clipped at zero. $E$ is fitted to three sets of points: the 400 held-out inputs, the decisions returned for the 100 objectives of the selection set, and the 2,000 points, out of 40,000 drawn uniformly from the variable box, with the smallest value of $|\hat h(x)+\hat q+c_{0}(x)|$,
where $c_{0}$ is an initial fit of the correction.

\paragraph{Matching cost.}

Since $E\ge 0$ by construction (it is a smooth maximum of affine functions and zero) and $F\ge 0$, the deployed sets are nested in $s$. Each method’s knob is placed on a grid on $\mathcal D_{\mathrm{match}}$ and then refined on the 200 evaluation objectives, by interpolation with a bisection fallback for at most 4 iterations (12 in one further pass on methane), until its cost is within $2\%$ of the budget.

\paragraph{Baselines.}

C-MICL deploys $\hat h(x)+k\,\hat q_{\mathrm{n}}\,\hat u(x)\le 0$, where
$\hat u$ is a network trained on the 1,000 training inputs to predict $|h-\hat h|$, $\hat q_{\mathrm{n}}$ is the conformal quantile of
$|h-\hat h|/\hat u$, and $k$ is the knob ($k=1$ is C-MICL as published). The C-MICL refit retrains $\hat u$ on the training inputs together with the 2,100
returned and boundary points of SRC's correction. W-MICL fits five surrogates and accepts a point when enough of them predict it feasible. The locally weighted margin scales its conformal quantile by the gap between the surrogate and a second, lower
quantile fit of $h$. Scalar padding adds a
constant to $\hat q$. $\alpha$-tightening recomputes $\hat q$ at a smaller $\alpha$. However, since $\hat q$ is a calibration residual, its cost moves in steps. The box trust region adds to $\hat h+\hat q\le 0$ the constraint that each variable lies between two quantiles of its training values, and its knob is the quantile level.

A comparison between a baseline and SRC's correction on one seed is made when both are matched, or when SRC's correction cost no more than the baseline.  On Hanoi at the $0.55$ budget, $\alpha$-tightening could not be compared on any seed.

{\scriptsize
\begin{longtable}{llccccc}
\caption{The convex-hull trust region of OptiCL against SRC's correction. For each of the
hull's three settings per seed, SRC's correction is taken at its most expensive setting
that costs no more than the hull. The $p$-value is from an exact McNemar test on the 200
shared objectives.}
\label{tab:opticl}\\
\toprule
 & & \multicolumn{2}{c}{convex hull} & \multicolumn{2}{c}{SRC correction} & \\
\cmidrule(lr){3-4}\cmidrule(lr){5-6}
benchmark & seed & cost & violations & cost & violations & $p$ \\
\midrule
\endfirsthead
\toprule
benchmark & seed & cost & violations & cost & violations & $p$ \\
\midrule
\endhead
Methane & 3 & 0.0508 & 17/200 & 0.0189 & 43/200 & 1.6e-04 \\
 &  & 0.0550 & 16/200 & 0.0189 & 43/200 & 6.6e-05 \\
 &  & 0.0597 & 9/200 & 0.0189 & 43/200 & 3.1e-07 \\
 & 5 & 0.0421 & 37/200 & 0.0379 & 0/200 & 1.5e-11 \\
 &  & 0.0468 & 34/200 & 0.0379 & 0/200 & 1.2e-10 \\
 &  & 0.0515 & 31/200 & 0.0379 & 0/200 & 9.3e-10 \\
 & 6 & 0.0438 & 35/200 & 0.0345 & 0/200 & 5.8e-11 \\
 &  & 0.0486 & 31/200 & 0.0345 & 0/200 & 9.3e-10 \\
 &  & 0.0532 & 28/200 & 0.0345 & 0/200 & 7.5e-09 \\
 & 7 & 0.0346 & 53/200 & 0.0338 & 0/200 & 2.2e-16 \\
 &  & 0.0388 & 43/200 & 0.0338 & 0/200 & 2.3e-13 \\
 &  & 0.0434 & 43/200 & 0.0400 & 0/200 & 2.3e-13 \\
 & 9 & 0.0377 & 63/200 & 0.0356 & 0/200 & 2.2e-19 \\
 &  & 0.0469 & 26/200 & 0.0356 & 0/200 & 3.0e-08 \\
 &  & 0.0568 & 11/200 & 0.0525 & 0/200 & 9.8e-04 \\
\addlinespace[3pt]
Williams--Otto & 3 & 0.0701 & 15/200 & 0.0259 & 3/200 & 0.002 \\
 &  & 0.0771 & 14/200 & 0.0259 & 3/200 & 0.007 \\
 &  & 0.0841 & 11/200 & 0.0259 & 3/200 & 0.039 \\
 & 5 & 0.0686 & 40/200 & 0.0314 & 0/200 & 1.8e-12 \\
 &  & 0.0756 & 40/200 & 0.0314 & 0/200 & 1.8e-12 \\
 &  & 0.0827 & 36/200 & 0.0314 & 0/200 & 2.9e-11 \\
 & 6 & 0.0685 & 24/200 & 0.0326 & 6/200 & 0.001 \\
 &  & 0.0756 & 26/200 & 0.0326 & 6/200 & 5.4e-04 \\
 &  & 0.0828 & 25/200 & 0.0326 & 6/200 & 8.8e-04 \\
 & 7 & 0.0619 & 19/200 & 0.0368 & 24/200 & 0.442 \\
 &  & 0.0693 & 18/200 & 0.0368 & 24/200 & 0.345 \\
 &  & 0.0766 & 15/200 & 0.0368 & 24/200 & 0.108 \\
 & 9 & 0.0724 & 34/200 & 0.0330 & 20/200 & 0.029 \\
 &  & 0.0797 & 23/200 & 0.0330 & 20/200 & 0.743 \\
 &  & 0.0870 & 21/200 & 0.0330 & 20/200 & 1.000 \\
\bottomrule
\end{longtable}}

\end{document}